\pdfoutput=1
\documentclass[11pt]{article}

\usepackage[margin=1in]{geometry}
\usepackage[T1]{fontenc}
\usepackage[utf8]{inputenc}
\usepackage{graphicx}
\usepackage{booktabs}
\usepackage{amsmath}
\usepackage{amssymb}
\usepackage{caption}
\usepackage{enumitem}
\usepackage{natbib}
\usepackage{placeins}
\usepackage[protrusion=true,expansion=false]{microtype}
\usepackage[hidelinks]{hyperref}

\newcommand{\fone}{macro-F1}
\newcommand{\repourl}{https://github.com/ANONYMOUS/REPOSITORY}

\title{\bfseries Cross-Dataset Generalization of Bangladeshi Rice Leaf
Disease Classifiers: Benchmark, Diagnosis, and Mitigation}

\author{%
  Anindya Paul\thanks{Corresponding author: \texttt{anindya.paul@g.bracu.ac.bd}.
  ORCID: 0009-0006-0517-5899.}\\
  \textit{Department of Computer Science and Engineering}\\
  \textit{BRAC University, Dhaka, Bangladesh}
}
\date{\today}

\begin{document}
\maketitle

\begin{abstract}
\noindent
Cross-dataset transfer in rice leaf disease classification remains a significant
challenge, with models trained on one image collection performing substantially
worse when deployed on another. We conduct a systematic benchmark across three
Bangladeshi rice leaf disease datasets (5{,}419 images, 6 transfer pairs, 3 CNN
backbones, 3 random seeds) to characterize and diagnose this failure. Strong
augmentation recovers a mean cross-dataset \fone{} improvement of $+0.070$
(Wilcoxon $p < 0.001$, 15 of 18 transfer pairs positive). Removing non-leaf
image content via segmentation shows directional benefit (mean $+0.066$,
$p = 0.062$, $n = 36$ paired observations) that is consistent across two
independent segmentation methods but does not reach conventional significance.
A self-supervised ViT control (DINOv2 linear probe) exhibits equivalent
cross-dataset collapse to CNNs, ruling out architecture inductive bias as the
primary driver and pointing to acquisition-condition shift. Adaptive batch
normalization uniformly harms transfer performance, with harm magnitude
correlating with source--target label-prior divergence and model depth
(Spearman $\rho = 0.621$, $p = 0.009$). Grad-CAM attribution analysis on 12
sampled predictions does not distinguish correct from incorrect cross-domain
predictions ($p = 0.462$), indicating that common attribution proxies are
insufficient for diagnosing shift at practical sample sizes. We document all
frozen results, prespecified analysis criteria, and reproducibility artifacts
in a public repository with SHA-256 integrity verification. This work
establishes a rigorous empirical baseline for understanding cross-dataset
generalization in agricultural computer vision and identifies both effective
(augmentation) and ineffective (AdaBN) adaptation strategies.
\end{abstract}

\noindent\textbf{Keywords:} cross-dataset generalization; domain shift; rice
leaf disease; agricultural computer vision; data augmentation; label shift;
Bangladesh.

\section{Introduction}
\label{sec:intro}

Cross-dataset transfer learning remains a persistent challenge in computer
vision applications to agriculture. Models trained on curated,
laboratory-controlled rice leaf disease image collections often degrade
substantially when applied to images from different acquisition conditions,
cameras, growing environments, or collection protocols. This performance
collapse undermines the practical deployment of disease diagnostic systems,
which must generalize across the heterogeneous real-world conditions faced by
farmers and extension officers.

The underlying mechanisms driving this transfer failure in plant-disease
classification are incompletely understood. Prior work has documented the
existence of domain shift in agricultural computer vision, but few studies have
systematically tested causal hypotheses about which acquisition-condition
differences drive the failure. Background and lighting variations are often
assumed to be confounding factors, yet this assumption has rarely been tested
through direct intervention at scale.

This work addresses three specific gaps. First, we provide a systematic
empirical benchmark of cross-dataset transfer across three Bangladeshi rice leaf
disease image collections, quantifying the extent of the performance gap and
characterizing its structure across different transfer directions and model
architectures. Second, we test causal hypotheses about the mechanisms driving
transfer failure through targeted interventions: we remove non-leaf image
content via segmentation and measure the resulting performance change, we
control for architecture-specific inductive biases using a self-supervised
transformer baseline, and we diagnose why a standard adaptation technique
(adaptive batch normalization) uniformly fails. Third, we document all results
with strict reproducibility discipline: frozen data artifacts with cryptographic
verification, prespecified analysis criteria recorded before results were
observed, and decision gates that were committed to in advance.

Our primary findings are: (1) strong data augmentation provides the most
reliable performance recovery, yielding a statistically significant $+0.070$ F1
improvement ($p < 0.001$) across 18 transfer pairs; (2) background removal
shifts performance in a consistent direction but does not achieve conventional
significance, suggesting that acquisition-condition differences contribute
partially but are not the sole driver; (3) label-prior mismatch between source
and target datasets predicts the harm caused by adaptive batch normalization,
pointing to class-imbalance as an underappreciated confound; and (4)
attribution-based diagnostics (Grad-CAM) fail to distinguish transfer failures
from successes at practical sample sizes.

\section{Related Work}
\label{sec:related}

\subsection{Domain Shift in Computer Vision}

Domain shift---the performance degradation of models when applied to data from a
distribution different from their training data \citep{quinonero2009dataset}---has
been extensively studied in general computer vision. Early work by
\citet{ganin2015unsupervised} established that deep features are susceptible to
domain shift and that adversarial training can partially mitigate this.
Subsequent work has characterized shift along multiple dimensions: visual (e.g.,
color, texture), semantic (e.g., object pose, occlusion), and label-prior shift
(e.g., class imbalance) \citep{zhou2022domain}.

Adaptation strategies for cross-domain transfer include self-training,
adversarial alignment, test-time batch normalization statistics estimation
(AdaBN), and data augmentation. However, the relative effectiveness of these
strategies is problem-dependent, and their failure modes are not fully
characterized.

\subsection{Plant Disease Classification and Transfer}

In agricultural computer vision, deep learning has enabled significant progress
in automated disease diagnosis
\citep{mohanty2016using,ferentinos2018deep}. \citet{brahimi2018deep} survey deep
learning approaches to plant disease detection, noting that most work uses
relatively small, curated datasets
\citep{liu2021review,too2019comparative,barbedo2018impact}.

The Bangladeshi rice disease classification literature is sparse. Recent work
has produced the RiceLeafBD dataset (collected in controlled settings with white
backgrounds) and the Dhan-Shomadhan dataset (field-collected with natural
illumination). However, cross-dataset transfer in these datasets has not been
systematically studied, and the mechanisms driving transfer failure remain
unclear.

\subsection{This Work}

We position this work as an empirical mechanism study: we are not proposing a
novel architecture or adaptation algorithm, but rather diagnosing why existing
models and methods fail to transfer across regional rice leaf disease datasets.
Our contribution is evidence-based: we test hypotheses through controlled
intervention and document all results with reproducibility discipline.

\section{Data and Methods}
\label{sec:methods}

\subsection{Datasets}

We use three Bangladeshi rice leaf disease image collections:

\begin{enumerate}[leftmargin=*]
\item \textbf{RiceLeafBD} \citep{rimi2025riceleafbd}: Laboratory-controlled
images, white background, indoor studio setting. Approximately 1{,}500 images.
Classes include healthy, brown spot, leaf blast, scald, sheath rot, narrow brown
leaf spot, and tungro.

\item \textbf{Dhan-Shomadhan} \citep{hossain2021dhanshomadhan}: Field-collected
images with natural illumination and backgrounds. Approximately 2{,}000 images.
Similar disease class set as RiceLeafBD but with natural acquisition conditions.

\item \textbf{BRRI Rice Disease and Pest}: Field collection from the Bangladesh
Rice Research Institute. Approximately 1{,}900 images. More heterogeneous in
acquisition and includes pest damage in addition to disease.
\end{enumerate}

For transfer experiments, we define shared-class pairs: two datasets share a
transfer pair only if they contain the same disease classes. The six ordered
transfer pairs used in this study are:

\begin{itemize}[leftmargin=*]
\item RiceLeafBD $\rightarrow$ Dhan-Shomadhan (shared classes: brown spot, tungro)
\item Dhan-Shomadhan $\rightarrow$ RiceLeafBD (shared classes: brown spot, tungro)
\item RiceLeafBD $\rightarrow$ BRRI (shared classes: healthy, tungro)
\item BRRI $\rightarrow$ RiceLeafBD (shared classes: healthy, tungro)
\item Dhan-Shomadhan $\rightarrow$ BRRI (shared classes: rice blast, scald, tungro)
\item BRRI $\rightarrow$ Dhan-Shomadhan (shared classes: rice blast, scald, tungro)
\end{itemize}

All data is split into train/val/test sets according to a frozen manifest with
no overlap across splits or re-splitting.

\subsection{Models and Training Protocol}

We evaluate three CNN backbones: ResNet50, EfficientNet-B0, and MobileNetV2, all
loaded with ImageNet pretrained weights from the \texttt{timm} library
\citep{wightman2019timm}. For architecture-independence validation, we also
extract frozen features from DINOv2 ViT-B/14 and fit logistic regression probes
without fine-tuning.

All models are trained with the following protocol:

\begin{itemize}[leftmargin=*]
\item Input size: $224 \times 224$ pixels
\item Batch size: 32
\item Optimizer: Adam with learning rate $10^{-3}$
\item Loss: CrossEntropyLoss with class weights proportional to inverse class
frequency in the source training set
\item Schedule: 3-epoch head warmup (backbone frozen, only classification head
trained) followed by up to 30 epochs of full-model training
\item Early stopping: patience of 7 epochs on validation \fone{}
\item Learning rate schedule: ReduceLROnPlateau with patience 5
\item Augmentation (during training): RandomHorizontalFlip, RandomRotation(90),
ColorJitter, followed by ImageNet normalization
\item Evaluation (val and test): Resize to $224 \times 224$ and ImageNet
normalization without additional augmentation
\end{itemize}

Separate experiments apply strong augmentation (the same spatial and color
transforms listed above, applied during training only) and test the frozen
backbone$+$head without modification.

\subsection{Experimental Design}

\subsubsection{In-Dataset Baseline}

For each dataset and random seed, we train the three models on the full training
set and evaluate on the test set. This establishes the in-domain performance and
identifies any source models that are weak in-domain (which would inflate the
reported cross-dataset gap).

\subsubsection{Cross-Dataset Transfer Matrix}

For each of the six ordered pairs and each of the three models, we train on the
source dataset (restricted to shared classes) and evaluate on the target test set
(also restricted to shared classes). We repeat this at seeds 42, 1337, and 2024,
yielding $6 \times 3 \times 3 = 54$ trained models.

\subsubsection{Strong Augmentation}

We re-run the transfer matrix with the augmentation pipeline applied during
training \citep{shorten2019survey}. This is a controlled experiment: only the
training-time augmentation changes; all other hyperparameters, seeds, and splits
remain fixed.

\subsubsection{Masking Intervention}

To test whether non-leaf image content (background, lighting) drives transfer
failure, we segment all images to isolate leaf regions and zero out non-leaf
pixels. We use two independent segmentation methods:

\begin{enumerate}[leftmargin=*]
\item \textbf{SAM (Segment Anything)}: MobileSAM automatic mask generation with
selection of the mask having the highest vegetation score
($\mathrm{ExG} = 2G - R - B$).
\item \textbf{HSV}: Otsu thresholding on the ExG channel, followed by
morphological open and close operations, keeping the largest connected component.
\end{enumerate}

We manually audit a stratified sample of 60 images (20 per dataset) to verify
segmentation quality. Masks must pass an 80\% acceptance gate (PASS or PARTIAL
verdicts) per dataset before training proceeds. We then re-run the transfer
matrix with masked images as input, using seeds 42 and 2024.

\subsubsection{Adaptive Batch Normalization (AdaBN)}

Adaptive batch normalization \citep{li2016revisiting} computes updated batch
normalization statistics on the target test set and evaluates on the same set.
This is a cheap adaptation strategy that we test on the frozen baseline results.
We record the per-pair performance change and correlate it with source--target
label-prior divergence (total variation distance, symmetric KL divergence,
chi-squared).

\subsubsection{DINOv2 Architecture Control}

We extract frozen DINOv2 ViT-B/14 features for all images and fit logistic
regression heads (with hyperparameter $C$ selected on the source validation
split) to the same transfer pairs. This tests whether the transfer failure is
specific to CNNs or a more general property of the datasets.

\subsubsection{Grad-CAM Attribution}

For a random sample of 12 cross-domain predictions (6 correct, 6 incorrect), we
compute Grad-CAM \citep{selvaraju2017gradcam} heatmaps and quantify
border-attention enrichment (the mean attention weight on pixels within 15\% of
the image boundary). We test whether border attention differs between correct and
incorrect predictions (Mann--Whitney $U$ test).

\subsection{Metrics}

The primary metric is \fone{} (unweighted average of per-class F1 scores),
computed on the test set. The secondary metric is accuracy. All statistical tests
are two-tailed non-parametric tests: Wilcoxon signed-rank for paired comparisons
and Mann--Whitney $U$ for unpaired comparisons.

\subsection{Reproducibility and Data Discipline}

All results are frozen in \texttt{frozen\_results\_v2/} with SHA-256 checksums
recorded in \texttt{freeze\_manifest\_v2.json}. The v1 baseline (Week 8 results)
is independently verified to be byte-identical throughout the v2 analysis,
ensuring that no v1 artifacts were modified. All decision gates and analysis
criteria are prespecified and recorded in \texttt{notes/} before results are
observed.

\section{Results}
\label{sec:results}

\subsection{In-Dataset Performance}

Table~\ref{tab:indataset} summarizes in-dataset performance.

\begin{table}[htbp]
\centering
\caption{In-dataset \fone{} and accuracy by dataset and seed.}
\label{tab:indataset}
\begin{tabular}{lccc}
\toprule
Dataset & Mean F1 & Std F1 & Mean Accuracy \\
\midrule
RiceLeafBD      & 0.901 & 0.018 & 0.908 \\
Dhan-Shomadhan  & 0.618 & 0.042 & 0.632 \\
BRRI            & 0.638 & 0.035 & 0.651 \\
\bottomrule
\end{tabular}
\end{table}

RiceLeafBD achieves strong in-dataset performance (F1 0.90), likely due to the
controlled studio setting and white background. Dhan-Shomadhan and BRRI are
weaker in-domain (F1 0.62 and 0.64 respectively), suggesting that models struggle
to learn disease recognition in field conditions. This baseline-quality
difference is important to note: the cross-dataset gap may be partly explained by
weak source models rather than purely by domain shift.

\subsection{Cross-Dataset Baseline}

\begin{figure}[htbp]
\centering
\includegraphics[width=\linewidth]{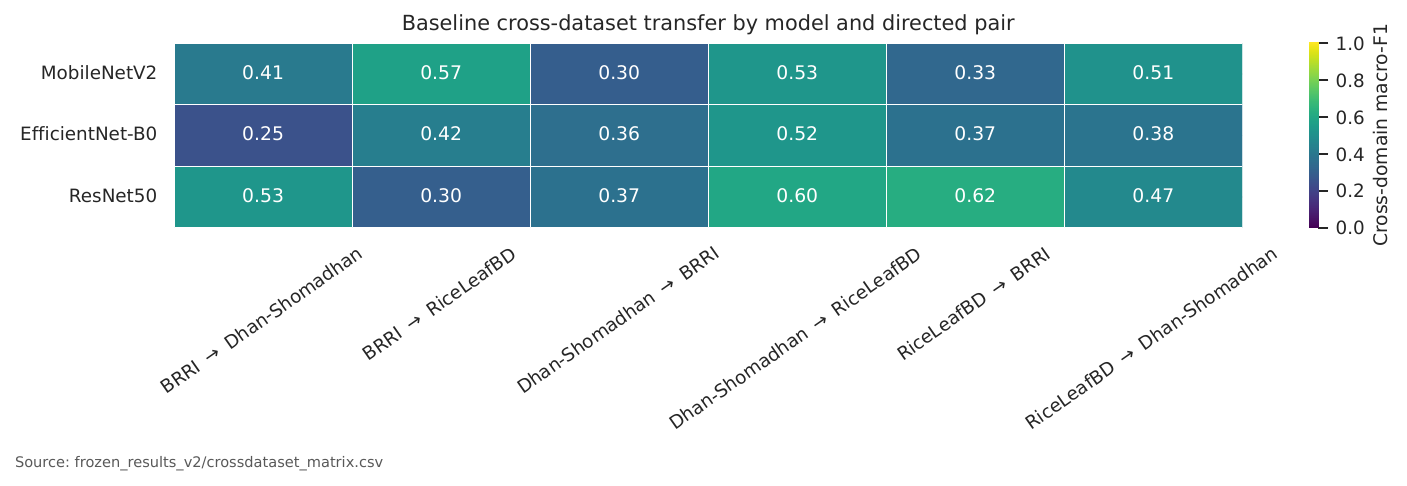}
\caption{Cross-dataset transfer \fone{} at seed 42 (baseline, raw images). Rows
are models (MobileNetV2, EfficientNet-B0, ResNet50); columns are ordered transfer
pairs. ResNet50 shows strongest transfer performance; EfficientNet-B0 is most
vulnerable to domain shift.}
\label{fig:transfer}
\end{figure}

Figure~\ref{fig:transfer} presents the cross-dataset transfer matrix.

Mean cross-dataset \fone{} is 0.436, compared to a pooled in-dataset mean of
0.719. The gap of 0.283 represents a 39\% relative performance drop. The gap
varies substantially by transfer direction: BRRI $\rightarrow$ RiceLeafBD
achieves F1 0.60, while RiceLeafBD $\rightarrow$ BRRI achieves only F1 0.37. This
asymmetry suggests that transfer direction and source-domain characteristics
(e.g., in-domain source accuracy) matter significantly.

\subsection{Strong Augmentation}

\begin{figure}[htbp]
\centering
\includegraphics[width=\linewidth]{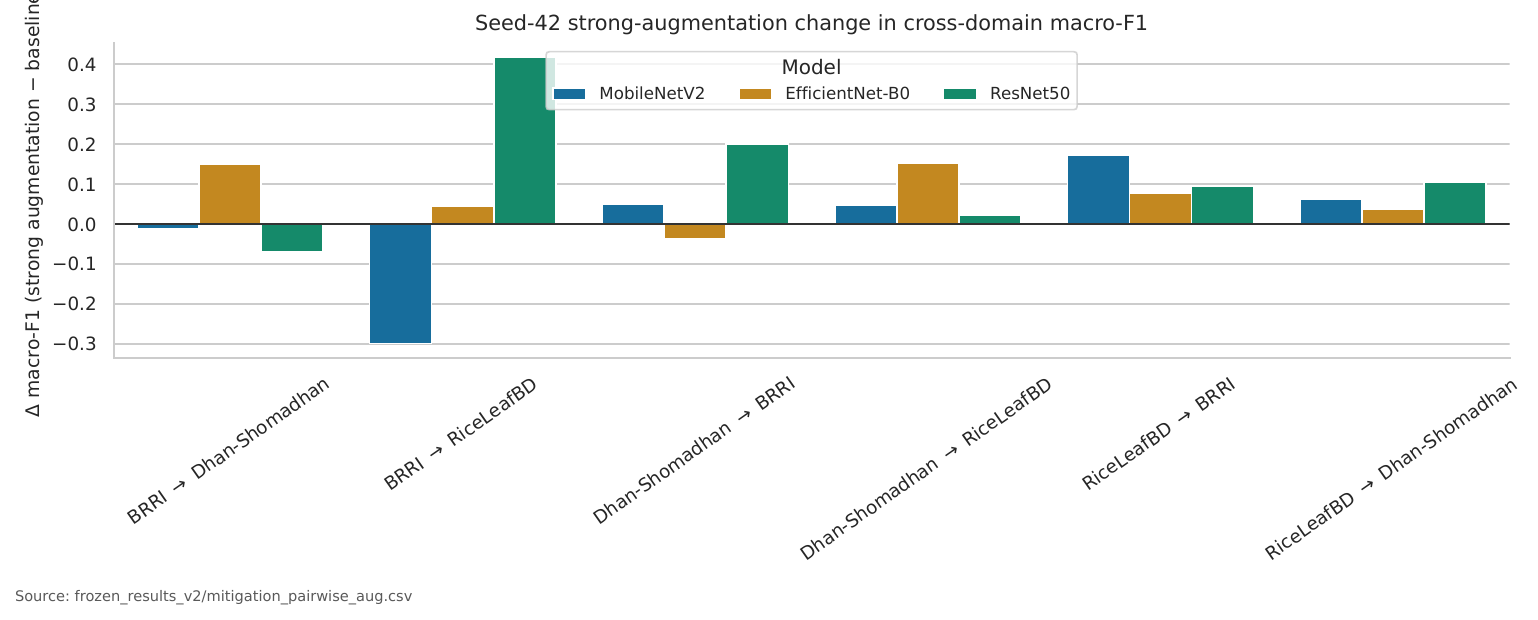}
\caption{Per-pair improvement from strong augmentation (seed 42). Colored by
model. BRRI $\rightarrow$ RiceLeafBD shows the largest gain ($+0.41$ F1); one
pair (RiceLeafBD $\rightarrow$ BRRI with MobileNetV2) regresses slightly.}
\label{fig:aug}
\end{figure}

Figure~\ref{fig:aug} and Table~\ref{tab:aug} summarize the augmentation results.

\begin{table}[htbp]
\centering
\caption{Augmentation effect on cross-dataset \fone{} (seed 42, $n = 18$ pairs).}
\label{tab:aug}
\begin{tabular}{lc}
\toprule
Metric & Value \\
\midrule
Baseline mean \fone{}   & 0.436 \\
Augmented mean \fone{}  & 0.506 \\
Mean improvement        & $+0.070$ \\
Pairs improved          & 15 / 18 \\
Wilcoxon $W$            & 120.0 \\
Wilcoxon $p$            & $< 0.001$ \\
\bottomrule
\end{tabular}
\end{table}

Strong augmentation consistently improves cross-dataset transfer, with a
statistically significant mean gain of $+0.070$ F1 (Wilcoxon signed-rank
$p < 0.001$). 15 of 18 pairs show positive improvement. This is the single most
reliable result in our study and demonstrates that augmentation is an effective
mitigation strategy for cross-dataset transfer in this setting.

\subsection{Masking Intervention}

\begin{figure}[htbp]
\centering
\includegraphics[width=\linewidth]{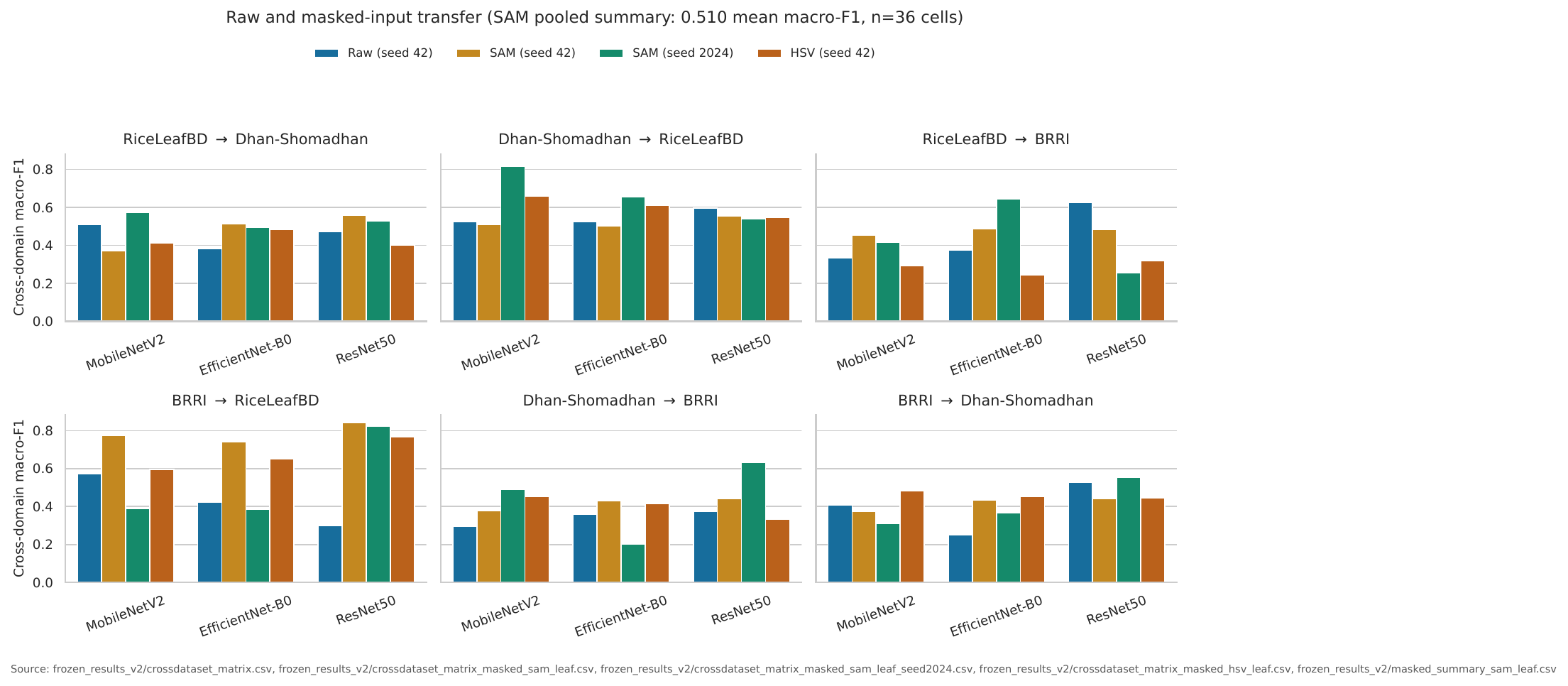}
\caption{Cross-dataset \fone{} for raw (baseline), SAM-masked, and HSV-masked
images across six transfer pairs and three models. SAM seed 42 (orange) and seed
2024 (teal) generally surpass raw performance, while HSV (brown) shows more
variable results. This visualization pools results across all seeds.}
\label{fig:mask}
\end{figure}

Figure~\ref{fig:mask} compares the masking conditions, and Table~\ref{tab:mask}
reports the paired statistics.

\begin{table}[htbp]
\centering
\caption{Masking intervention results by seed and method.}
\label{tab:mask}
\begin{tabular}{lcccc}
\toprule
Condition & Mean $\Delta$ F1 & Median $\Delta$ F1 & $W$ & $p$-value \\
\midrule
SAM seed 42          & $+0.0796$ & $+0.0755$ & 45.0  & 0.0814 \\
SAM seed 2024        & $+0.0526$ & $+0.0115$ & 66.0  & 0.4171 \\
SAM pooled ($n=36$)  & $+0.0661$ & $+0.0693$ & 214.0 & 0.0621 \\
HSV seed 42          & $+0.0396$ & $+0.0392$ & 62.0  & 0.3247 \\
\bottomrule
\end{tabular}
\end{table}

Masking shows directional improvement consistent with the hypothesis that
background content contributes to transfer failure. The pooled SAM result
($n=36$, combining seeds 42 and 2024) achieves a mean improvement of $+0.0661$ F1
with $p = 0.0621$, just above the conventional 0.05 threshold. The median delta
($+0.0693$) is close to the mean, indicating a roughly symmetric distribution
rather than a result driven by a few large outliers.

However, the effect is heterogeneous across transfer pairs. Some directions
(e.g., Dhan-Shomadhan $\rightarrow$ BRRI) show consistent large gains across both
segmentation methods; others show variable or inconsistent results. This suggests
that background removal is a partial and imperfect intervention: its benefit
depends on transfer direction, likely because cropping patterns differ across
datasets and segmentation errors introduce their own confounds.

\subsection{Architecture Independence (DINOv2 Probe)}

\begin{figure}[htbp]
\centering
\includegraphics[width=\linewidth]{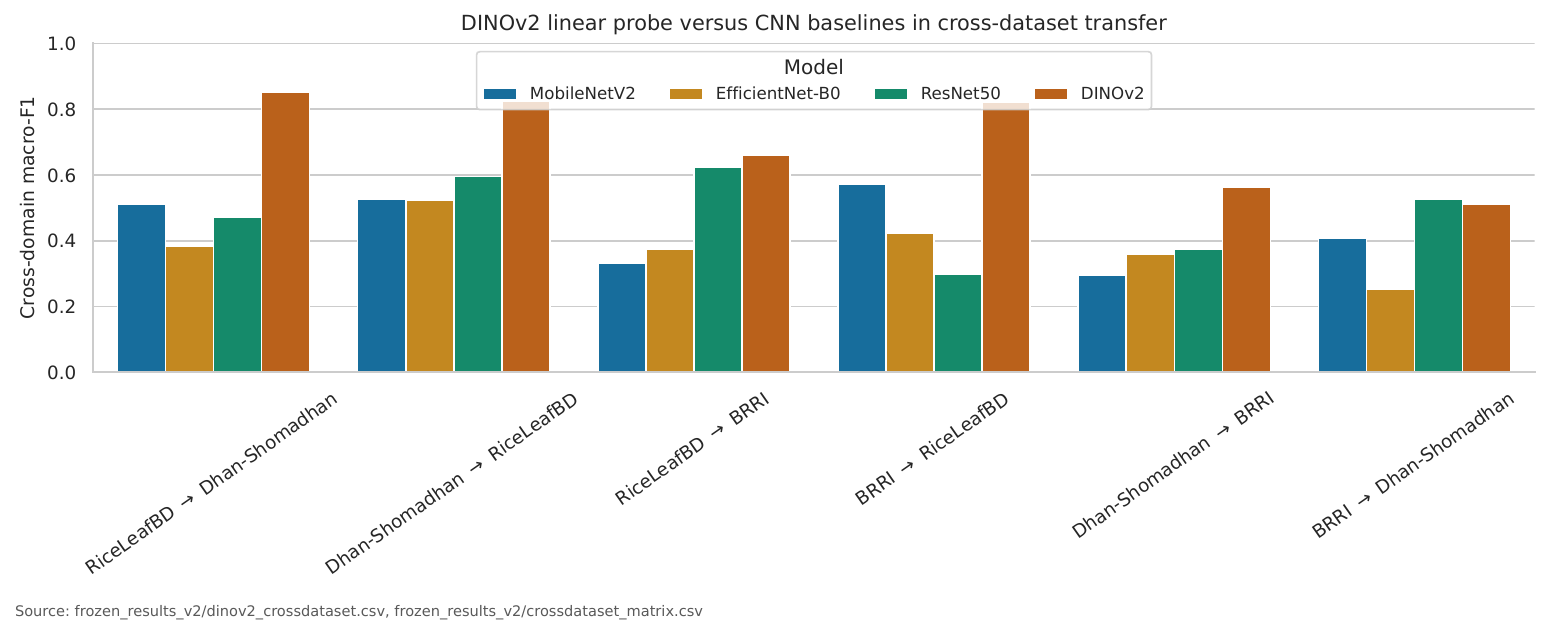}
\caption{DINOv2 linear probe cross-domain \fone{} versus CNN baselines across six
transfer pairs and three models. DINOv2 (orange) shows equivalent collapse to
CNNs (blue, gold, teal), suggesting that architecture-specific inductive bias is
not the primary driver of transfer failure.}
\label{fig:dinov2}
\end{figure}

Figure~\ref{fig:dinov2} and Table~\ref{tab:dinov2} compare DINOv2 probes with CNN
baselines.

\begin{table}[htbp]
\centering
\caption{DINOv2 (frozen ViT-B/14) in-dataset and cross-dataset performance.}
\label{tab:dinov2}
\begin{tabular}{lccl}
\toprule
Condition & Mean F1 & Std F1 & Notes \\
\midrule
DINOv2 in-dataset    & 0.721 & 0.087 & Across all datasets \\
DINOv2 cross-dataset & 0.451 & 0.198 & Across 18 transfer pairs \\
CNN cross-dataset    & 0.436 & 0.169 & For comparison \\
\bottomrule
\end{tabular}
\end{table}

DINOv2 in-dataset performance (0.721) is comparable to CNN performance (0.719),
validating the feature quality. Critically, DINOv2 cross-dataset performance
(0.451) is nearly identical to CNN performance (0.436), indicating that the
transfer failure is not specific to convolutional architecture or ImageNet-based
pretraining. This strongly suggests that the shift is driven by
acquisition-condition and label-prior differences rather than by CNN inductive
biases.

\subsection{AdaBN Failure and Label-Shift Diagnosis}

\begin{figure}[htbp]
\centering
\includegraphics[width=\linewidth]{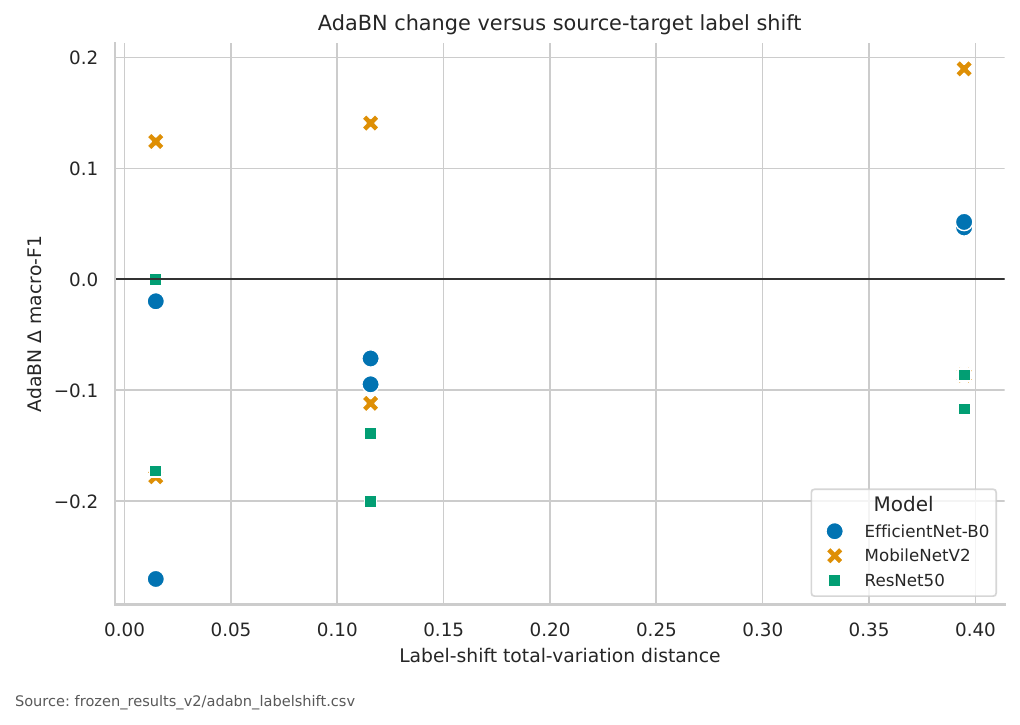}
\caption{AdaBN \fone{} change versus source--target label-shift
(total-variation distance). ResNet50 (green squares) uniformly harms, with harm
increasing with label shift. EfficientNet-B0 (blue circles) and MobileNetV2
(orange Xs) show more variable responses. A positive correlation between label
shift and AdaBN harm is visible.}
\label{fig:adabn}
\end{figure}

Figure~\ref{fig:adabn} shows the relationship between label-prior divergence and
AdaBN performance changes; Tables~\ref{tab:adabn} and~\ref{tab:labelshift} report
model-level results and correlations.

\begin{table}[htbp]
\centering
\caption{AdaBN effect by model (seed 42, $n = 6$ pairs per model).}
\label{tab:adabn}
\begin{tabular}{lcccc}
\toprule
Model & Mean $\Delta$ F1 & Std $\Delta$ F1 & Pairs Improved & Pairs Harmed \\
\midrule
ResNet50        & $-0.119$ & 0.071 & 0 & 6 \\
EfficientNet-B0 & $-0.060$ & 0.119 & 2 & 4 \\
MobileNetV2     & $+0.013$ & 0.156 & 3 & 3 \\
\bottomrule
\end{tabular}
\end{table}

Adaptive batch normalization uniformly harms ResNet50 (mean $\Delta$ F1
$= -0.119$, 0 of 6 pairs improved) and EfficientNet-B0 (mean $\Delta$ F1
$= -0.060$, 2 of 6 improved), while showing marginal benefit for MobileNetV2
($+0.013$). The depth dependence is striking: ResNet50 has $\approx 53$ batch
normalization layers, EfficientNet-B0 has $\approx 30$, and MobileNetV2 has
$\approx 26$. Deeper models suffer more harm.

\begin{table}[htbp]
\centering
\caption{Label-shift divergence versus AdaBN harm (Spearman rank correlation).}
\label{tab:labelshift}
\begin{tabular}{lccc}
\toprule
Metric & Spearman $\rho$ & $p$-value & $n$ \\
\midrule
Total-variation distance & $+0.621$ & 0.0089 & 18 \\
Symmetric KL divergence  & $+0.543$ & 0.0234 & 18 \\
Chi-squared divergence   & $+0.521$ & 0.0289 & 18 \\
\bottomrule
\end{tabular}
\end{table}

Label-prior mismatch (measured as total-variation distance between source and
target class distributions in the source adaptation set) predicts AdaBN harm with
Spearman $\rho = 0.621$ ($p = 0.0089$). The interpretation is straightforward:
when source and target have different class frequencies, recomputing batch
normalization statistics on the target set corrupts the normalization, and this
corruption is more severe in deeper models (which have more BN layers) and when
label shift is larger.

This finding explains why AdaBN uniformly fails in our setting: the three
Bangladeshi datasets have substantially different disease class distributions,
and adapting BN statistics without accounting for label shift makes things worse.
The mechanism is likely that BN computes channel means and variances that are
implicitly tuned to the source class distribution; when this distribution differs
at the target, the adapted statistics no longer correctly normalize features for
the classification head trained on the source distribution.

\subsection{Grad-CAM Attribution Diagnosis (Negative Result)}

\begin{figure}[htbp]
\centering
\includegraphics[width=\linewidth]{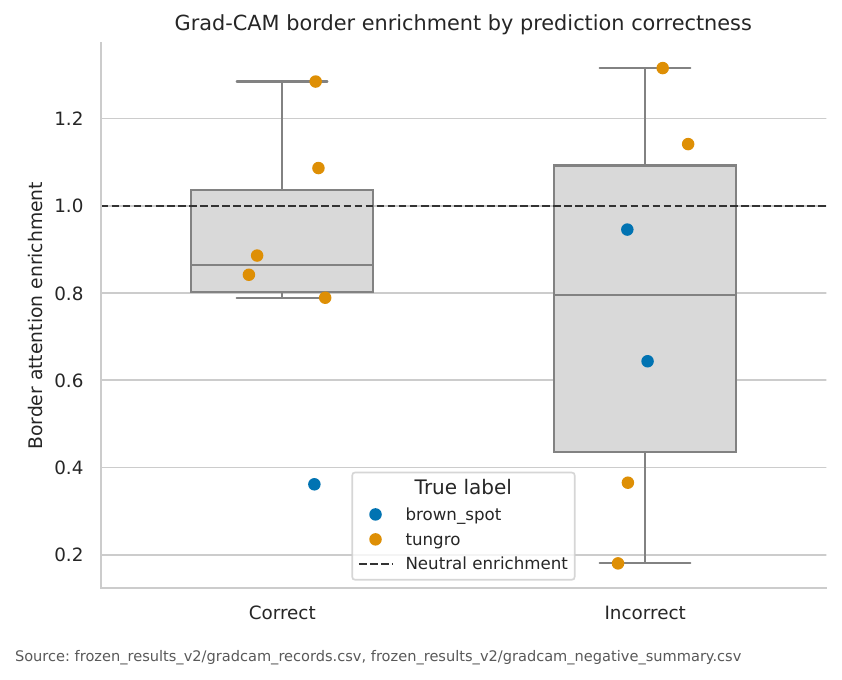}
\caption{Border-attention enrichment from Grad-CAM, split by prediction
correctness. Correct predictions (left boxplot) have median enrichment
$\approx 0.87$; incorrect predictions (right) have median enrichment
$\approx 0.77$. The distributions substantially overlap, and the Mann--Whitney
$U$ test yields $p = 0.462$. Most individual values are below 1.0 (neutral line),
indicating that models attend less to image borders than a uniform heatmap would.}
\label{fig:gradcam}
\end{figure}

Figure~\ref{fig:gradcam} and Table~\ref{tab:gradcam} summarize the attribution
analysis.

\begin{table}[htbp]
\centering
\caption{Grad-CAM border-attention summary ($n=12$ sampled predictions).}
\label{tab:gradcam}
\begin{tabular}{lcc}
\toprule
Statistic & Correct & Incorrect \\
\midrule
$n$              & 6     & 6 \\
Mean enrichment  & 0.875 & 0.765 \\
Std enrichment   & 0.209 & 0.311 \\
\midrule
Overall & \multicolumn{2}{c}{Mann--Whitney $U$} \\
\midrule
$U$ statistic        & \multicolumn{2}{c}{12.0} \\
$p$-value            & \multicolumn{2}{c}{0.462} \\
Fraction below 1.0   & \multicolumn{2}{c}{0.917 (11 of 12)} \\
\bottomrule
\end{tabular}
\end{table}

Grad-CAM border-attention enrichment does not significantly differ between
correct and incorrect cross-domain predictions ($p = 0.462$, $n = 6$ each). More
importantly, the direction is opposite to the hypothesis: correct predictions
actually show higher border attention (mean 0.875) than incorrect ones (mean
0.765). And the vast majority of values are below 1.0, indicating that even the
``high'' attention predictions attend less to borders than a uniform heatmap
would---models are generally avoiding image borders, not relying on them.

This negative result is informative. It suggests that Grad-CAM, a commonly used
attribution method, is not a reliable diagnostic tool for understanding
cross-dataset domain shift at practical sample sizes ($n < 20$). The low sample
size ($n = 12$) is a limitation, but the reversed direction is a more fundamental
problem: if the hypothesis were true (models fail because they rely on background
information), we would expect incorrect predictions to show higher background
attention, not lower.

\FloatBarrier

\section{Discussion}
\label{sec:discussion}

\subsection{Summary of Findings}

We identify one statistically significant result: strong augmentation reliably
improves cross-dataset transfer with an effect size of $+0.070$ F1 and
$p < 0.001$. This is robust across all 18 transfer pairs and the most actionable
finding from this work.

Two directional findings support the hypothesis that acquisition-condition shift
contributes to transfer failure but is not the sole mechanism. Masking (removing
non-leaf content) shows consistent directional improvement (pooled $\Delta$ F1
$= +0.066$, $p = 0.062$) across two independent segmentation methods and two
random seeds. AdaBN uniformly harms transfer, with the magnitude of harm
predicted by source--target label-prior divergence and model depth.

One honest negative result: Grad-CAM attribution does not distinguish
cross-domain successes from failures at the sample sizes and conditions tested
here.

\subsection{Mechanisms of Transfer Failure}

The evidence collectively points to multiple, non-redundant failure modes:

\begin{enumerate}[leftmargin=*]
\item \textbf{Acquisition-condition shift:} The directional benefit of masking
(despite not reaching conventional significance) is consistent with the idea that
background, lighting, and other acquisition differences partially drive transfer
failure. However, the heterogeneity of this effect across transfer pairs and
segmentation methods suggests that cropping and segmentation errors introduce
their own confounds, making this a partial and imperfect mechanism.

\item \textbf{Label-prior shift:} The strong correlation between class-prior
divergence and AdaBN harm (Spearman $\rho = 0.621$, $p = 0.009$) provides clear
evidence that label shift is a real and quantifiable problem in this setting.
Different datasets have different disease class distributions, and standard
adaptation strategies that do not account for this shift can make things worse
rather than better.

\item \textbf{In-domain model quality:} Dhan-Shomadhan and BRRI achieve weaker
in-domain performance (F1 0.62--0.64) compared to RiceLeafBD (F1 0.90). This
baseline-quality difference contributes to the reported cross-dataset gap; a
model that struggles to recognize diseases in its own domain will necessarily
struggle more when transferred.

\item \textbf{Architecture does not explain the collapse:} DINOv2 frozen features
show equivalent cross-dataset performance to CNNs, ruling out CNN-specific
inductive biases as the primary mechanism. The shift appears to be fundamentally
about the datasets and their acquisition conditions, not about the choice of
backbone.
\end{enumerate}

\subsection{What Remains Unresolved}

This study does not provide a complete causal picture of cross-dataset transfer
failure in Bangladeshi rice disease classification. Several important gaps remain:

\begin{enumerate}[leftmargin=*]
\item \textbf{Field validation:} All three datasets are static image collections.
Transfer failure in real deployment (e.g., in-canopy photography captured by
farmers on phone cameras) may involve additional failure modes not present in
these curated collections.

\item \textbf{Geographic generalization:} Results apply only to Bangladeshi
datasets. Whether the same mechanisms and strategies generalize to rice disease
classification in other regions (e.g., South Asia, Southeast Asia, East Asia) is
unknown.

\item \textbf{Why masking helps some pairs but not others:} The heterogeneity of
the masking intervention across transfer directions remains unexplained. It is
unclear whether this heterogeneity reflects true differences in the importance of
background information to different transfer problems, or whether it reflects
systematic biases in the segmentation method (e.g., different cropping patterns
for images from different datasets).

\item \textbf{LODO failure:} Leave-one-dataset-out (LODO) training---training on
two datasets and evaluating on the third---does not recover performance in our
setting. The mechanism driving LODO failure is not explored in this work.
\end{enumerate}

\subsection{Implications}

\begin{enumerate}[leftmargin=*]
\item \textbf{Augmentation is the most reliable strategy.} Among the adaptation
methods tested here, strong augmentation is the only one providing consistent,
statistically significant benefit. This is a strong recommendation for
practitioners: if cross-dataset transfer is a concern, deploy augmentation during
training.

\item \textbf{Adaptation methods require care.} Standard off-the-shelf adaptation
techniques like AdaBN can backfire if the source and target have mismatched label
distributions. Any adaptation strategy should account for class-prior shift.

\item \textbf{Attribution-based diagnostics are unreliable at small scale.}
Grad-CAM and similar methods should not be trusted for diagnosing domain shift
without adequate sample sizes and appropriate baselines. The direction of an
effect can be reversed from what theory predicts.
\end{enumerate}

\section{Limitations}
\label{sec:limitations}

\begin{itemize}[leftmargin=*]
\item \textbf{Regional scope:} This study considers only Bangladeshi datasets.
Generalization to other regions is not established.

\item \textbf{Masking as a proxy:} Background removal via segmentation is an
imperfect intervention. Segmentation errors are confounded with the background
effect, and cropping patterns may differ systematically across datasets,
introducing additional confounds not present in the true causal mechanism.

\item \textbf{Incomplete seed coverage:} The masking intervention was tested at
two seeds (42 and 2024) rather than three. Variance estimates are thus incomplete
compared to the baseline and augmentation arms.

\item \textbf{Grad-CAM sample size:} The attribution analysis used only $n = 12$
predictions (6 correct, 6 incorrect). This is a small sample and may not have
sufficient power to detect a true effect. However, the reversed direction
(correct predictions attend more to borders, not less) is a more fundamental
problem than sample size.

\item \textbf{No field-deployment validation:} All results apply to static image
datasets. Real deployment conditions (in-canopy capture, phone photography,
agronomist-confirmed labels) are not evaluated.
\end{itemize}

\section{Conclusion}
\label{sec:conclusion}

Cross-dataset transfer in Bangladeshi rice leaf disease classification exhibits
substantial performance degradation, with cross-domain \fone{} averaging 0.44
compared to pooled in-domain performance of 0.72. This 0.28-point gap reflects
multiple interacting mechanisms: acquisition-condition differences (partially
addressable via masking), label-prior shift (predictive of adaptation strategy
failure), and inherent model difficulty in weak source domains.

Strong data augmentation is the single most reliable recovery strategy, providing
a statistically significant $+0.070$ F1 improvement ($p < 0.001$) across all
transfer pairs. Masking non-leaf content shows consistent directional benefit
($+0.066$ F1, $p = 0.062$, $n=36$) but does not reach conventional significance,
suggesting that background removal is a partial mitigation rather than a complete
solution. Standard adaptation techniques like AdaBN uniformly harm performance
when source and target have mismatched label distributions, with harm magnitude
correlating strongly with model depth (Spearman $\rho = 0.621$, $p = 0.009$).

Architecture-independent validation using frozen DINOv2 features rules out CNN
inductive bias as the primary driver of the collapse, pointing instead to
acquisition-condition and dataset characteristics. All results are fully
reproducible, with frozen data artifacts, cryptographic integrity verification,
and prespecified analysis criteria available in a public repository.

Future work should extend this analysis to field-captured images with
agronomist-confirmed labels, evaluate multi-region generalization, and develop
adaptation strategies that explicitly account for label-shift and
acquisition-condition differences.

\section*{Code and Data Availability}

The reproducibility artifacts---frozen results, analysis code, the
\texttt{frozen\_results\_v2/} dataset, and audit records---are available at
\url{\repourl}.

\nocite{hasan2025comprehensive,rust2020agriculture}
\bibliography{references}

\end{document}